\documentclass[conference]{IEEEtran}
\usepackage{times}

\usepackage{amsmath,amsfonts,bm}

\def\Figref#1{Fig.~\ref{#1}}

\def\eqref#1{equation~\ref{#1}}
\def\Eqref#1{Eq.~\ref{#1}}

\def\1{\bm{1}}

\def\rmU{{\mathbf{U}}}

\DeclareMathAlphabet{\mathsfit}{\encodingdefault}{\sfdefault}{m}{sl}
\SetMathAlphabet{\mathsfit}{bold}{\encodingdefault}{\sfdefault}{bx}{n}

\newcommand{\E}{\mathbb{E}}

\DeclareMathOperator*{\argmax}{arg\,max}

\usepackage[numbers]{natbib}
\usepackage{multicol}
\usepackage[bookmarks=true]{hyperref}
\usepackage{graphicx}
\usepackage{amsmath}
\usepackage{amssymb}
\usepackage{amsfonts}
\usepackage{mathtools, xparse}
\usepackage{algorithm}
\usepackage{algorithmicx}
\usepackage{algpseudocode}
\usepackage{multirow}
\usepackage[normalem]{ulem}
\useunder{\uline}{\ul}{}
\usepackage{hyperref}
\hypersetup{
    colorlinks=true,
    linkcolor=blue,
    filecolor=magenta,      
    urlcolor=blue,
}
\usepackage[font={small}]{caption} 
\renewcommand{\baselinestretch}{0.995} 

\newcommand{\xxnote}[3]{}
\ifx\hidenotes\undefined
  \renewcommand{\xxnote}[3]{\color{#2}{#1: #3}}
\fi

\begin{document}

\title{Learning to Manipulate Deformable Objects \\without Demonstrations}


\author{
Yilin Wu$^*$, Wilson Yan$^*$, Thanard Kurutach, Lerrel Pinto, Pieter Abbeel \\
University of California, Berkeley\\
\texttt{\{yilin-wu,wilson1.yan\}@berkeley.edu}
\thanks{$^*$Equal contribution, authors are listed in alphabetical order}
}%


%

\maketitle

\begin{abstract}
In this paper we tackle the problem of deformable object manipulation through model-free visual reinforcement learning (RL). In order to circumvent the sample inefficiency of RL, we propose two key ideas that accelerate learning. First, we propose an iterative pick-place action space that encodes the conditional relationship between picking and placing on deformable objects. The explicit structural encoding enables faster learning under complex object dynamics. Second, instead of jointly learning both the pick and the place locations, we only explicitly learn the placing policy conditioned on random pick points. Then, by selecting the pick point that has Maximal Value under Placing (MVP), we obtain our picking policy. This provides us with an informed picking policy during testing, while using only random pick points during training. Experimentally, this learning framework obtains an order of magnitude faster learning compared to independent action-spaces on our suite of deformable object manipulation tasks with visual RGB observations. Finally, using domain randomization, we transfer our policies to a real PR2 robot for challenging cloth and rope coverage tasks, and demonstrate significant improvements over standard RL techniques on average coverage.  
\end{abstract}
\IEEEpeerreviewmaketitle

\section{Introduction}

Over the last few decades, we have seen tremendous progress in robotic manipulation. From grasping objects in clutter~\cite{shimoga1996robot,pinto2016supersizing,mahler2016dexnet,levine2016learning,gupta2018robot} to dexterous in-hand manipulation of objects~\cite{andrychowicz2018learning,yousef2011tactile}, modern robotic algorithms have transformed object manipulation. But much of this success has come at the price of making a key assumption: rigidity of objects. Most robot algorithms often require (implicitly or explicitly) strict rigidity constraints on objects. But the objects we interact with everyday, from the clothes we put on to shopping bags we pack, are deformable. In fact, even `rigid' objects deform under different form factors (like a metal wire). Because of this departure from the `rigid-body' assumption, several real-world applications of manipulation fail~\cite{johnson2015team}. So why haven't we created equally powerful algorithms for deformable objects yet?



\begin{figure}
  \begin{center}
    \includegraphics[width = \linewidth]{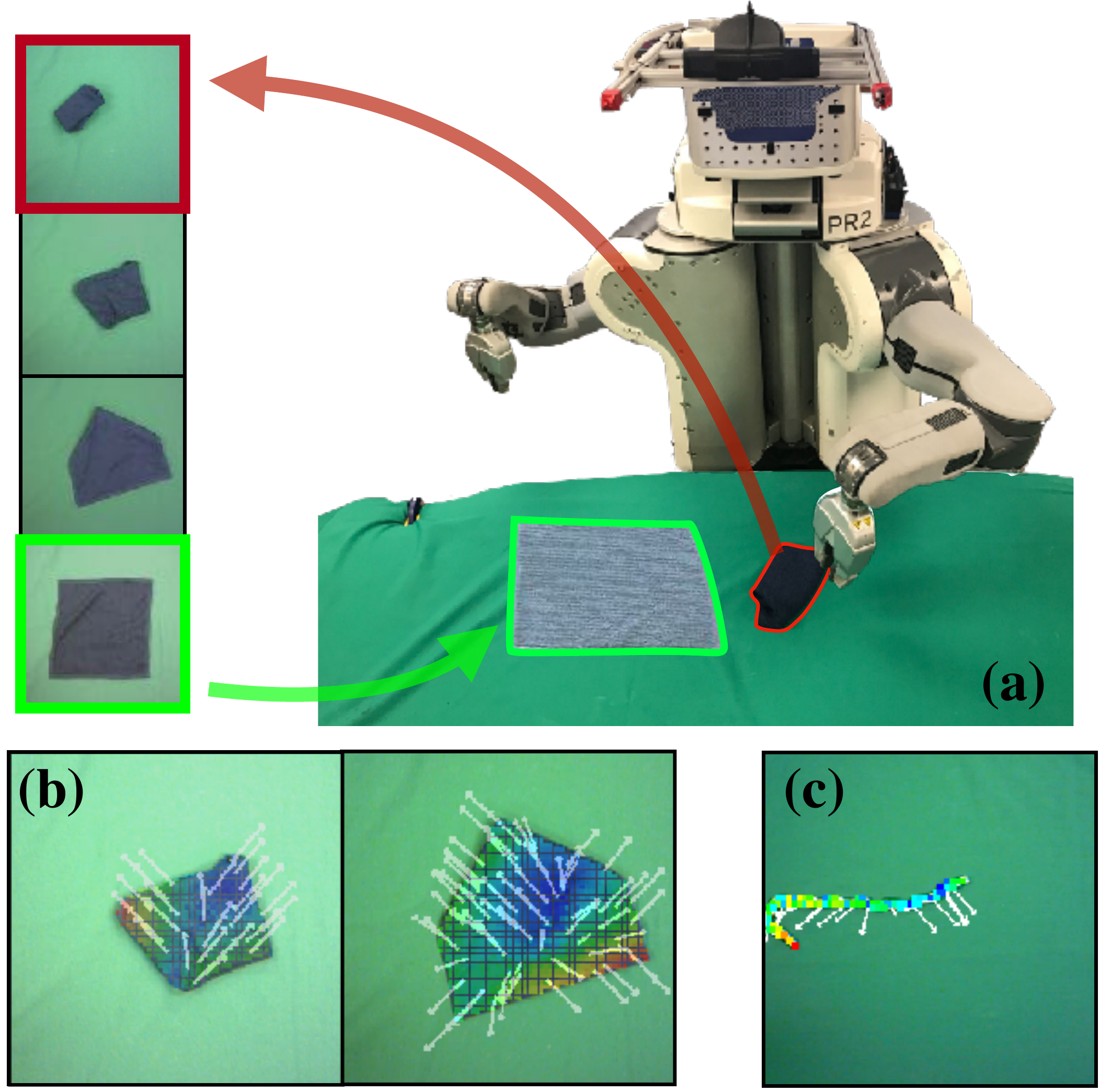}
  \end{center}
  \caption{ We look at the problem of deformable object manipulation, where the robot needs to manipulate a deformable object, say the blue cloth, into a desired goal location (green in (a)). Our method learns an explicit placing policy (arrows in (b) and (c)), along with an implicit picking policy. This method is evaluated on cloth (b) and rope (c) tasks using our PR2 robot. The heatmaps represent the distribution of the Q-value, where the Q-values over each pick location are normalized to the range of 0 (blue) to 1 (red).}
\label{fig:intro}
\end{figure}

Deformable object manipulation has been a long standing problem~\cite{wada2001robust,henrich2012robot,stria2014garment,maitin2010cloth,seita2018deep}, with two unique challenges. First, in contrast with rigid objects, there is no obvious representation of state. Consider the cloth manipulation problem in \Figref{fig:intro}(a), where the robot needs to flatten a cloth from any start configuration. How do we track the shape of the cloth? Should we use a raw point cloud, or fit a continuous function? This lack of canonical state often limits state representations to discrete approximations~\cite{berenson2013manipulation}. Second, the dynamics is complex and non-linear~\cite{essahbi2012soft}. Due to microscopic interactions in the object, even simple looking objects can exhibit complex and unpredictable behavior~\cite{pieranski2001tight}. This makes it difficult to model and perform traditional task and motion planning. 

One of the recent breakthroughs in robotics has been the development of model-free visual policy learning~\cite{levine2016end,pinto2017asymmetric,andrychowicz2018learning}, where robotic algorithms can reason about interactions directly from raw sensory observations. This can alleviate the challenge of state estimation for deformable objects~\cite{matas2018sim}, since we can directly learn on images. Moreover, since these methods do not require an explicit model of the object~\cite{lillicrap2015continuous}, they can overcome the challenge of having complex deformable object dynamics. But model-free learning has notoriously poor sample complexity~\cite{duan2016benchmarking}. This has limited the application of learning to the setting where human demonstrations are available~\cite{nair2017combining,matas2018sim}. To reduce the dependence on human demonstrators, \citet{seita2019deep} in concurrent and independent work, has shown how expert state-based policies can provide simulated demonstrations to learn cloth manipulation from visual observations.



In this work, we tackle the sample-complexity issue by focusing on an often ignored aspect of learning: the action space. Inspired by \citet{howard2000intelligent,brooks1983planning}, we start by using an iterative pick-place action space, where the robot can decide which point to grasp (or pick) and to which point it should drop (or place). But how should one learn with this action space? One option is to directly output both the pick point and place location for the deformable object. But the optimal placing location is heavily correlated with picking location, i.e. where you place depends heavily on what point you pick. This conditional structure makes it difficult to simultaneously learn without modeling the action space.


To solve this, we propose a conditional action space, where the output of the picking policy is fed as input into the placing policy. This type of action space is inspired by recent work in auto-regressive output spaces in image generation~\cite{van2016conditional}, imitation learning~\cite{nair2017combining}, and grasping~\cite{tobin2018domain}. However in the context of model-free RL, this leads us to a second problem: the placing policy is constrained by the picking policy. When learning starts, the picking policy often collapses into a suboptimal restrictive set of pick points. This inhibits the exploration of the placing policy, since the picking points it takes as input are only from a restrictive set, and results in a suboptimal placing policy. Now, since the rewards for picking come after the placing is executed, the picking policy receives poor rewards and results in inefficient learning. This illustrates the chicken and egg problem with conditional action spaces. Learning a good picking strategy involves having a good placing strategy, while learning a good placing strategy involves having a good picking strategy.

To break this chicken and egg loop, we learn the placing strategy independent of the picking strategy. This allows us to both learn the placing policy efficiently, and use the learned placing value approximator~\cite{lillicrap2015continuous} to inform the picking policy. More concretely, since the value of the placing policy is conditioned on the pick point, we can find the pick point that maximizes the value. We call this picking policy Maximum Value of Placing (MVP). During training, the placing policy is trained with a random picking policy. However, during testing, the MVP picking policy is used. Through this, we observe a significant speedup in convergence on three difficult deformable object manipulation tasks on rope and cloth objects. Finally, we demonstrate how this policy can be transferred from a simulator to a real robot using simple domain randomization without any additional real-world training or human demonstrations. Videos of our PR2 robot performing deformable object
manipulation along with our code can be accessed on the project website: \url{https://sites.google.com/view/alternating-pick-and-place}. Interestingly, our policies are able to generalize to a variety of starting states of both cloth and rope previously unseen in training.



In summary, we present three contributions in this paper: 
 (a) we propose a novel learning algorithm for picking based on the maximal value of placing; (b) we show that the conditional action space formulation significantly accelerates the learning for deformable object manipulation; and (c) we demonstrate transfer to real-robot cloth and rope manipulation using our proposed formulation.



\section{Related Work}
\subsection{Deformable Object Manipulation}
Robotic manipulation of deformable objects has had a rich history that has spanned different fields from surgical robotics to industrial manipulation. For a more detailed survey, we refer the reader to \citet{khalil2010dexterous,henrich2012robot}.

Motion planning has been a popular approach to tackle this problem, where several works combine deformable object simulations with efficient planning~\cite{jimenez2012survey}. Early work~\cite{saha2007manipulation,wakamatsu2006knotting,moll2006path} focused on using planning for linearly deformable objects like ropes. \citet{rodriguez2006obstacle} developed methods for fully deformable simulation environments, while \citet{frank2011efficient} created methods for faster planning with deformable environments. One of the challenges of planning with deformable objects is the large degrees of freedom and hence large configuration space involved when planning. This, coupled with the complex dynamics~\cite{essahbi2012soft}, has prompted work in using high-level planners or demonstrations and local controllers to follow the plans.

Instead of planning on the full complex dynamics, we can plan on simpler approximations, but use local controllers to handle the actual complex dynamics. One way to use local controllers is model-based servoing~\cite{smolen2009deformation,wada2001robust}, where the end-effector is locally controlled to a given goal location instead of explicit planning. However, since the controllers are optimized over simpler dynamics, they often get stuck in local minima with more complex dynamics~\cite{mcconachie2017interleaving}. To solve this model-based dependency, ~\citet{berenson2013manipulation,mcconachie2018estimating,navarro2014visual} have looked at Jacobian approximated controllers that do not need explicit models, while \citet{jia2018learning,hu2018three} have looked at learning-based techniques for servoing.  However, since the controllers are still local in nature, they are still susceptible to reaching globally suboptimal policies. To address this, \citet{mcconachie2017interleaving} interleaves planning along with local controllers. Although this produces better behavior, transferring it to a robot involves solving the difficult state-estimation problem~\cite{schulman2013generalization,schulman2013tracking}. Instead of a two step planner and local controller, we propose to directly use model-free visual learning, which should alleviate the state-estimation problem along with working with the true complex dynamics of the manipulated objects.

\subsection{Reinforcement Learning for Manipulation}
Reinforcement Learning (RL) has made significant progress in many areas including robotics. RL has enabled robots to handle unstructured perception such as visual inputs and reason about actions directly from raw observations \cite{mnih2015human}, which can be desirable in many robotic tasks. RL from vision has been shown to solve manipulation problems such as in-hand block manipulation \cite{andrychowicz2018learning, rajeswaran2017learning}, object pushing \cite{finn2017deep}, and valve-rotating with a three-fingered hand \cite{haarnoja2018soft}. However, these algorithms have not yet seen wide applicability to deformable object manipulation. This is primarily due to learning being inefficient with complex dynamics~\cite{duan2016benchmarking}, which we address in this work.

Over the last few years, deformable object manipulation has also been studied in reinforcement learning \cite{nair2017combining, lee2015learning, matas2018sim, wang2019learning, seita2019deep}. However, many of these works~\cite{lee2015learning, matas2018sim} require expert demonstrations to guide learning for cloth manipulation. These expert demonstrations can also be used to learn wire threading ~\cite{mayer2008system, schulman2013case}. In concurrent work, ~\citet{seita2019deep} show that instead of human demonstrators, a simulated demonstrator using state information can be used to obtain demonstrations. Other works like \citet{nair2017combining} that do not need demonstrations for training require them at test time. Similarly, \citet{wang2019learning} do not use demonstration, but do require self-supervised exploration data at training time. We note that since using our conditional action spaces and MVP technique can be applied to any actor-critic algorithm, it is complementary to most methods that learn from expert demonstrations.




\section{Background}

\begin{figure*}
  \begin{center}
    \includegraphics[width = \textwidth]{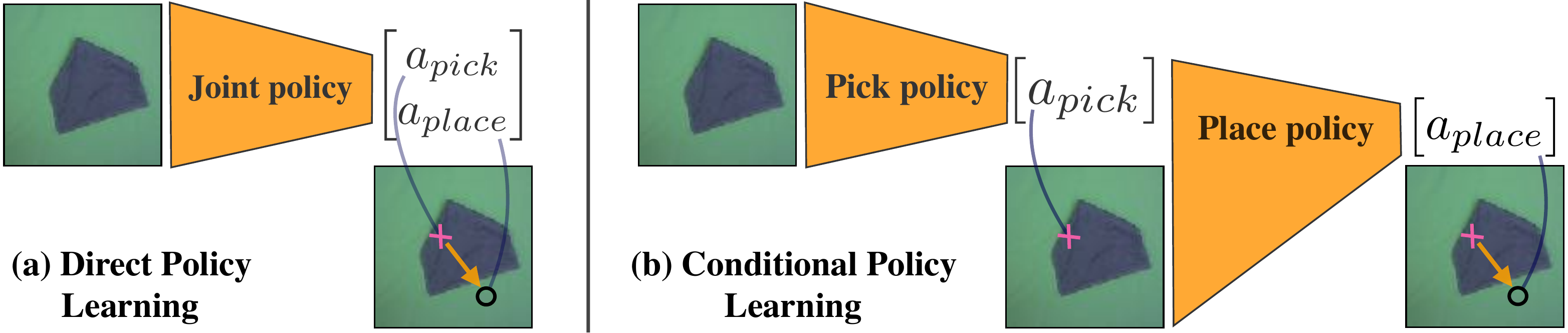}
  \end{center}
  \caption{In direct policy learning (a), the policy directly outputs both the pick and the place location. While in conditional policy learning, the composite action space is broken down into a separate picking and placing policy, where the placing policy takes the output of the picking policy as input.}
\label{fig:arch}
\end{figure*}

Before we describe our learning framework, we briefly discuss relevant background on reinforcement learning and off-policy learning. For a more in-depth survey, we refer the reader to~\citet{sutton1998introduction,kaelbling1996reinforcement}.

\subsection{Reinforcement Learning}
We consider a continuous Markov Decision Process (MDP), represented by the tuple \begin{math}(\mathcal{S},\mathcal{O},\mathcal{A},\mathcal{P},r,\mathcal{\gamma},s_0)\end{math}, with continuous state and action space, $\mathcal{S}$ and $\mathcal{A}$, and a partial observation space $\mathcal{O}$. \begin{math}\mathcal{P : S \times A \times S} {\rightarrow} [0,\infty)\end{math} defines the transition probability of the next state \begin{math}s_{t+1}\end{math} given the current state-action pair \begin{math}(s_t,a_t)\end{math}. For each transition, the environment generates a reward  \begin{math}r : \mathcal{S \times A \rightarrow R}\end{math}, with future reward discounted by $\gamma$. 

Starting from an initial state $s_0$ sampled from distribution $\mathcal{S}$, the agent takes actions according to policy \begin{math}\pi(a_t|s_t)\end{math} and receives reward $r_t = r(s_t,a_t)$ at every timestep t. The next state $s_{t+1}$ is sampled from the transition distribution \begin{math}\mathcal{P}(s_{t+1}|s_t,a_t)\end{math}. The objective in reinforcement learning is to learn a policy that maximizes the expected sum of discounted rewards \begin{math}\sum_t\E_{(s_t,a_t) \sim \rho_{\pi}(s_t,a_t)}[\gamma^t r(s_t,a_t)]\end{math}. In the case of a partially observable model, the agent receives observations $o_t$ and learns $\pi(a_t|o_t)$.
\subsection{Off Policy Learning}

On-policy reinforcement learning ~\cite{schulman2015trust,kakade2002natural,williams1992simple} iterates between data collection and policy updates, hence requiring new on-policy data per iteration which tends to be expensive to obtain. On the other hand, off-policy reinforcement learning retains past experiences in a replay buffer and is able to re-use past samples. Thus, in practice, off-policy algorithms have achieved significantly better sample efficiency \cite{haarnoja2018soft, kurutach2018model}.
Off-policy learning can be divided into three main categories: model-based RL, Actor-Critic (AC), and Q learning. In model-based RL, we learn the dynamics of the system. In the AC framework, we learn both the policy (actor) and value function (critic). Finally, in Q-learning we often learn only the value function, and choose actions that maximize it.

In this work, we choose the actor-critic framework due to its stability, data-efficiency, and suitability for continuous control.
Recent state-of-the-art actor-critic algorithms such as Twin Delayed DDPG (TD3)~\cite{2018arXiv180209477F}  and Soft-Actor-Critic (SAC)~\cite{2018arXiv180101290H} show better performance than prior off-policy algorithms such Deep Deterministic Policy Gradient (DDPG)~\cite{2015arXiv150902971L} and Asynchronous Advantage Actor-Critic (A3C)~\cite{2016arXiv160201783M} due to variance reduction methods in TD3 by using a second critic network to reduce over-estimation of the value function and an additional entropy term in SAC to encourage exploration. In this work, we use SAC since its empirical performance surpasses TD3 (and other off-policy algorithms) on most RL benchmark environments~\cite{2018arXiv180101290H}. However, our method is not tied to SAC and can work with any off-policy learning algorithm. 

\subsection{Soft Actor Critic}
For our experiments, we use SAC~\cite{haarnoja2018soft}, an entropy regularized off-policy RL algorithm, as our base RL algorithm. This regularization allows for a trade-off between the entropy of the policy and its expected return. Intuitively, increasing the entropy makes the policy more exploratory, which helps prevent convergence to poor solutions.

SAC learns a parameterized Q-function $Q_\theta(s,a)$ and policy $\pi_\phi(a|s)$, and an entropy-regularization term $\alpha$. $Q_\theta$ is learned by minimizing a bootstrapped estimate of the Q-value using a target Q-value network with an included entropy term. $\pi_\phi$ is learned by minimizing the expected KL-divergence between $\phi_\phi(\cdot|s_t)$ and $\frac{\exp\{Q_\theta(s_t,\cdot)\}}{Z_\theta(s_t)}$, where $Z_\theta(s_t)$ is a normalization constant. Lastly, the entropy regularization term $\alpha$ is learned  by iteratively adjusting to fit a target entropy. We choose to use SAC over existing off-policy methods since it has shown consistently better results in many popular domains. 

\section{Approach}

We now describe our learning framework for efficient deformable object manipulation. We start by the pick and place problem. Following this, we discuss our algorithm.


\subsection{Deformable Object Manipulation as a Pick and Place Problem}

We look at a more amenable action space while retaining the expressivity of the general action space: pick and place. The pick and place action space has had a rich history in planning with rigid objects~\cite{brooks1983planning, lozano1989task}. Here, the action space is the location to pick (or grasp) the object $a^t_{pick}$
and the location to place (or drop) the object $a^t_{place}$. This operation is done at every step $t$, but we will drop the superscript for ease of reading. With rigid objects, the whole object hence moves according $a_{pick}\rightarrow a_{place}$. However, for a deformable object, only the point corresponding to $a_{pick}$ on the object moves to $a_{place}$, while the other points move according to the kinematics and dynamics of the deformable object \cite{nair2017combining}. Empirically, since in each action the robot picks and places a part of the deformable object, there is significant motion in the object, which means that the robot gets a more informative reward signal after each action. Also note that this setting allows for multiple pick-and-place operations that are necessary for tasks such as spreading out a scrunched up piece of cloth.


\subsection{Learning with Composite Action Spaces}
The straightforward approach to learning with a pick-place action space is to learn a policy $\pi_{joint}$ that directly outputs the optimal locations to pick and to place $\lbrack a_{pick}, a_{place} \rbrack$, i.e. $\pi_{joint} \equiv p(a_{pick}|o)\cdot p(a_{place} | o)$ where $o$ is the observation of the deformable object (\Figref{fig:arch}(a)). However, this approach fails to capture the underlying composite and conditional nature of the action space, where the location to place $a_{place}$ is strongly dependent on the pick point $a_{pick}$. 

One way to learn with conditional output spaces is to explicitly factor the output space during learning. This has provided benefits in several other learning problems from generating images~\cite{van2016conditional} to predicting large dimensional robotic actions~\cite{murali2018cassl,tobin2018domain}. Hence instead of learning the joint policy, we factor the policy as:

\begin{equation}
    \pi_{factor} \equiv \pi_{pick}(a_{pick}|o) \cdot \pi_{place}(a_{place}|o,a_{pick})
\end{equation}

This factorization will allow the policy to reason about the conditional dependence of placing on picking (\Figref{fig:arch}(b)). However, in the context of RL, we face another challenge: action credit assignment. Using RL, the reward for a specific behavior comes through the cumulative discounted reward at the end of an episode. This results in the \textit{temporal credit assignment} problem where attributing the reward to a specific action is difficult. With our factored action spaces, we now have an additional credit assignment problem on the different factors of the action space. This means that if an action receives high reward, we do not know if it is due to $\pi_{pick}$ or $\pi_{place}$. Due to this, training $\pi_{factor}$ jointly is inefficient and often leads to the policy selecting a suboptimal pick location. This suboptimal $\pi_{pick}$ then does not allow $\pi_{place}$ to learn, since $\pi_{place}(a_{place}|o,a_{pick})$ only sees suboptimal picking locations $a_{pick}$ during early parts of training. Thus, this leads to a mode collapse as shown in Sec. \ref{sec:fact_benefits}.

To overcome the action credit assignment problem, we propose a two-stage learning scheme. 
Here the key insight is that training a placing policy can be done given a full-support picking policy and the picking policy can be obtained from the placing policy by accessing the Value approximator for placing.
Algorithmically, this is done by first training $\pi_{place}$ conditioned on picking actions from the uniform random distribution $\rmU_{pick}$. Using SAC, we train and obtain $\pi_{place}(a_{place}|o,a_{pick})$, $\text{s.t. } a_{pick} \sim \rmU_{pick}$ as well as the place value approximator $V^{\pi_{place}}_{place}(o,a_{pick})$. Since the value is also conditioned on pick point $a_{pick}$, we can use this to obtain our picking policy as:

\begin{equation}
\label{eq:mvp}
    \pi_{pick} \equiv \argmax_{a_{pick}} V^{\pi_{place}}_{place}(o,a_{pick})
\end{equation}

We call this picking policy: Maximum Value under Placing (MVP). The $\argmax$ is computed by searching over all available pick location from the image of the object being manipulated. MVP allows us get an informed picking policy without having to explicitly train for picking. This makes training efficient for off-policy learning with conditional action spaces especially in the context of deformable object manipulation. 
\section{Experimental Evaluation}

In this section we analyze our method MVP across a suite of simulations and then demonstrate real-world deformable object manipulation using our learned policies.

\begin{figure*}
  \begin{center}
    \includegraphics[width = \textwidth]{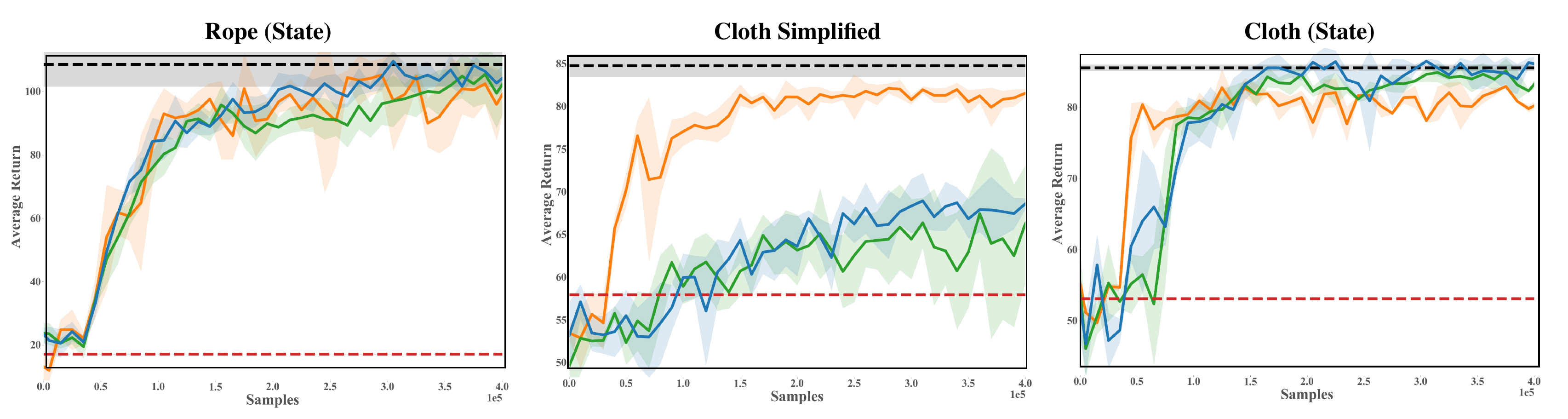}
  \end{center}
  \caption{Learning comparisons between baselines and our method on the three deformable object manipulation environments with state-based training in simulation. The dotted black line is computed by evaluating MVP on the final learned `learned placing with uniform pick' policy. Each experiment was run on 4 random seeds.}
\label{fig:sim_plots_state}
\end{figure*}

\begin{figure*}
  \begin{center}
    \includegraphics[width = 0.8\textwidth]{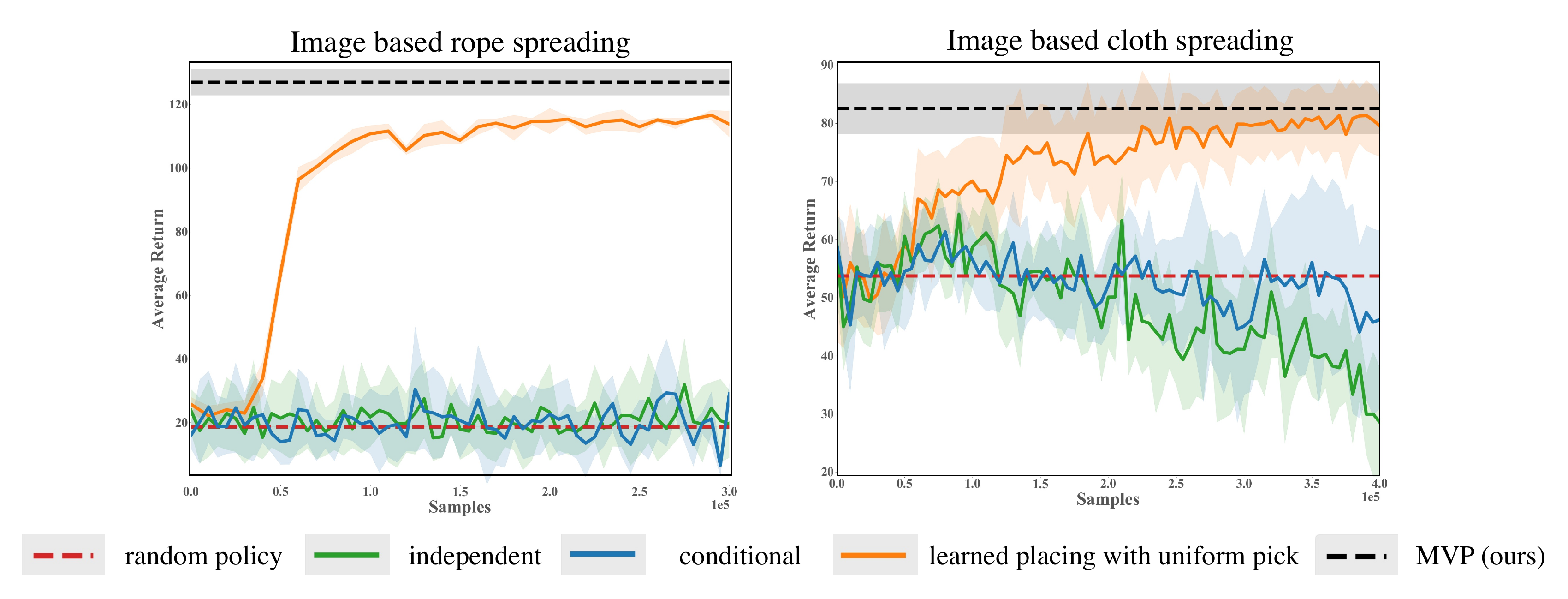}
  \end{center}
  \caption{Learning comparisons between baselines and our method on two deformable object manipulation environments with image-based training in simulation. Note that we do not include the \textit{cloth-simplified} environment here since image-based transfer to real robot would involve corner detection. The dotted black line is computed by evaluating MVP on the final learned `learned placing with uniform pick' policy. Each experiment was run on 3 random seeds.}
\label{fig:sim_plots_image}
\end{figure*}

\subsection{Cloth Manipulation in Simulation}
Most current RL environments like OpenAI Gym~\cite{brockman2016openai} and DM Control~\cite{tassa2018deepmind}, offer a variety of rigid body manipulation tasks. However, they do not have environments for deformable objects. Therefore, for consistent analysis, we build our own simulated environments for deformable objects using the DM Control API. To simulate deformable objects, we use composite objects from MuJoCo 2.0~\cite{todorov2012mujoco}. This allows us to create and render complex deformable objects like cloths and ropes. Using MVP, we train policies both on state (locations of the composite objects) and image observations ($64\times64\times3$ RGB). For image-based experiments, we uniformly randomly select a pick point on a binary segmentation of the cloth or rope in order to guarantee a pick point on the corresponding object. Images are segmented using simple color channel thresholding. The details for the three environments we use are as follows:

\begin{figure*}[ht!]
  \begin{center}
    \includegraphics[width = 0.9\linewidth]{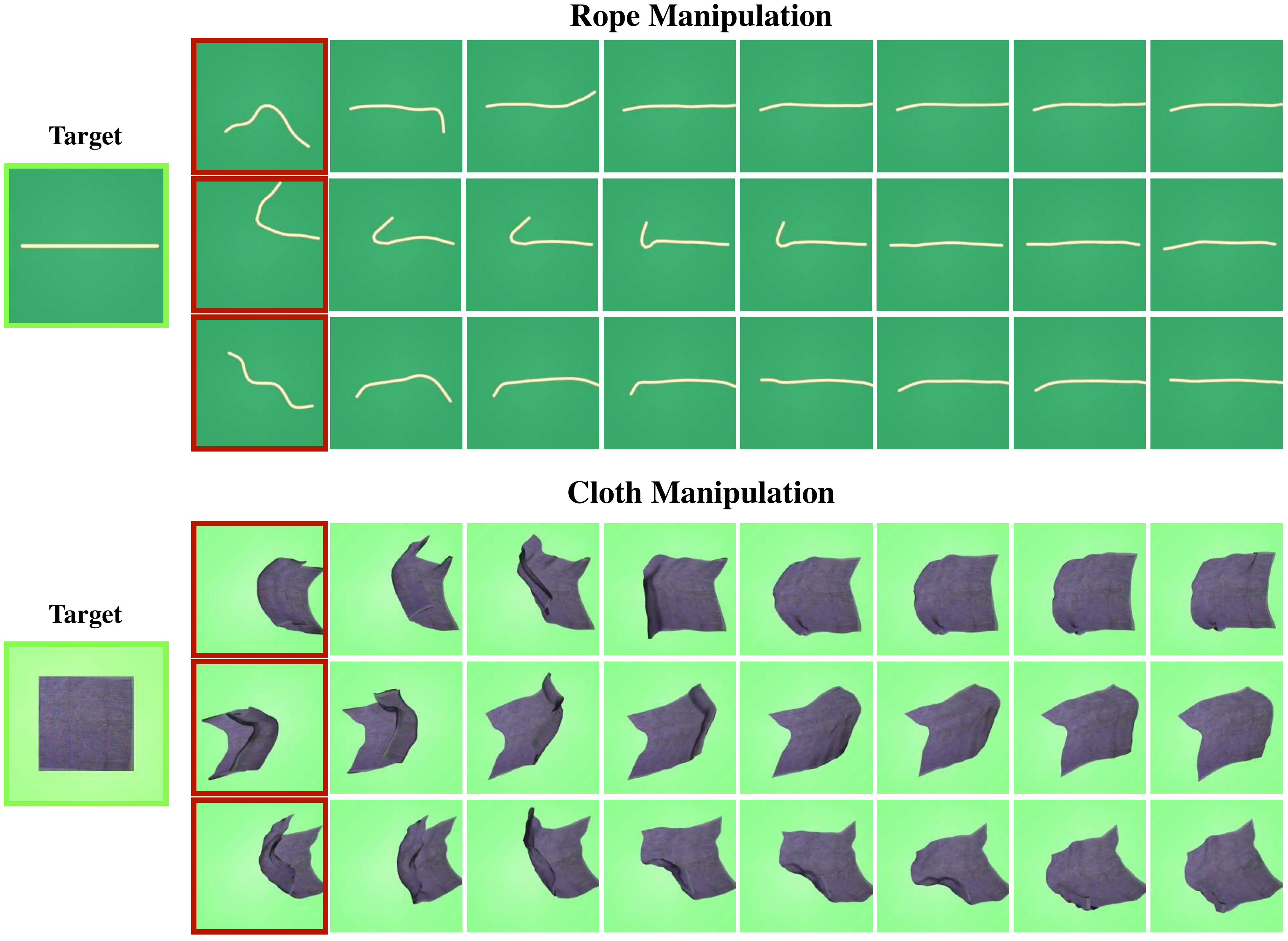}
  \end{center}
  \caption{We demonstrate deformable object manipulation in the simulated environments using our learned MVP policy. In the top half, we see the policy successfully horizontally straightens and centers a rope in the top. And in the bottom half, we see our method successfully spreading out a cloth from multiple starting states. Each image is about 5 actions apart for rope experiments, and 10 actions for cloth experiments.}
\label{fig:sim_exp}
\end{figure*}

\begin{figure*}[ht!]
  \begin{center}
    \includegraphics[width = 0.9\linewidth]{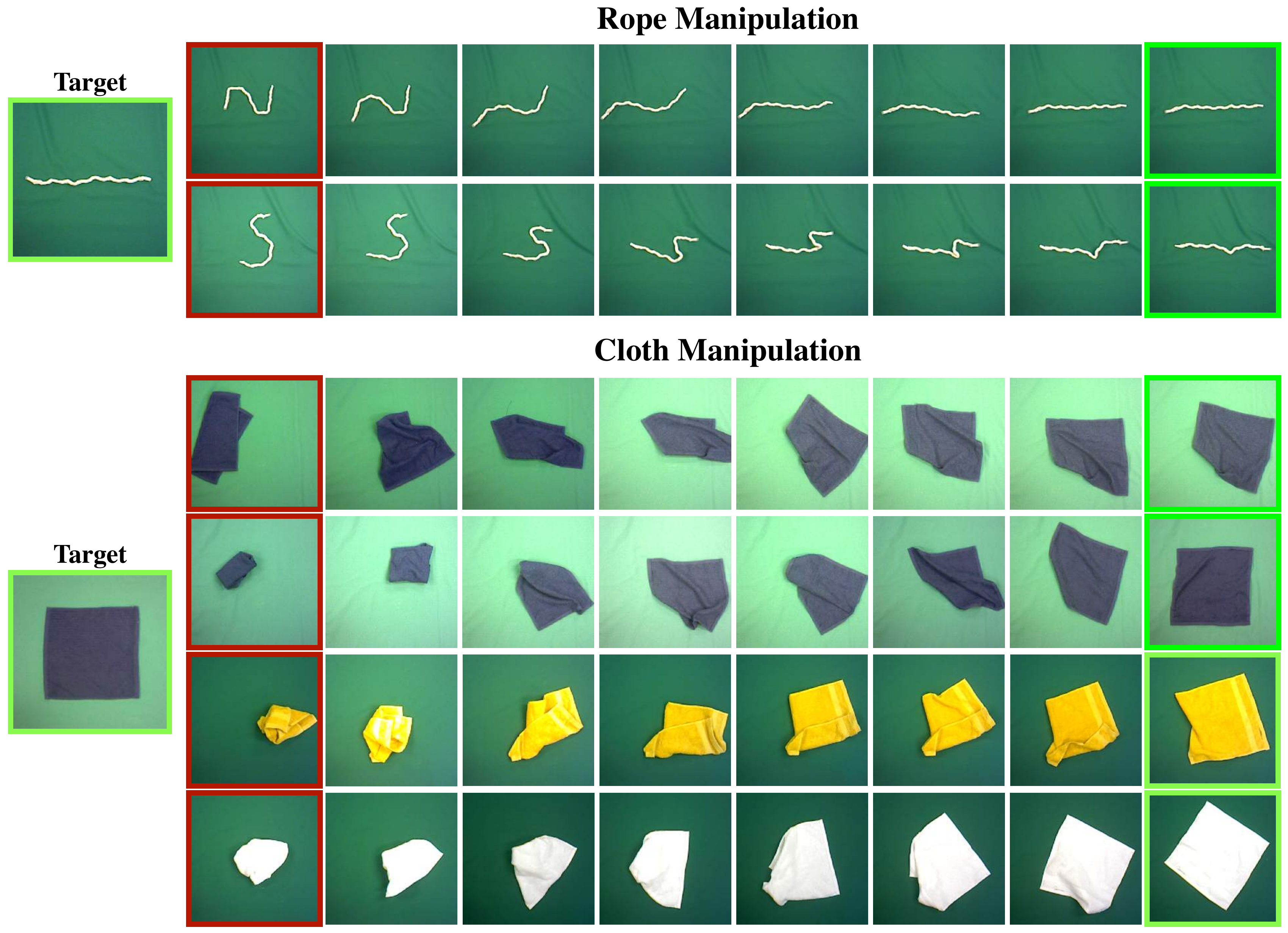}
  \end{center}
  \caption{Using MVP for learning the policy along with domain randomization for simulation to real transfer, we demonstrate deformable object manipulation on a real robot for a rope (top) and a cloth (bottom). In both examples, the task is to spread out the object to reach the target spread out configuration in the middle of the table (left) for two different start locations (in red). For rope spreading, each frame corresponds to one pick-place action taken by our PR2 robot (\Figref{fig:intro}(a)), while for cloth spreading each frame corresponds to 10 actions on our robot.}
\label{fig:robot_exp}
\end{figure*}



  \par \textbf{1. Rope} : The goal is to stretch the rope (simulated as a 25 joint composite) horizontally straight in the center of the table. The action space is divided into two parts as $a_{pick}$ and $a_{place}$. $a_{pick}$ is the two dimension pick point on the rope, and $a_{place}$ is the relative distance to move and place the rope. All other parts of the rope move based on the simulator dynamics after each action is applied. We constrain the relative distance to move in a small radius around the pick point due to unstable simulations for larger movements. The reward for this task is computed from the segmentation of the rope in RGB images as: 
  
  \begin{equation}
   \begin{split}
     reward = \sum_{i=1}^{H}e^{0.5\times|i-32|} \sum_{j=1}^{W} s_{i,j} 
\end{split},
  \end{equation}
  where $i$ is the row number of the image, $j$ is the column number, $s_{i,j}$ is the binary segmentation at pixel location $(i,j)$, and $W, H$ correspond to height and width. Hence for a $64\times 64$ image the reward encourages the rope to be in the center row (row number $32$) with an exponential penalty on rows further from the center. At the start of each episode, the rope is initialized by applying a random action for the first 50 timesteps.

  \par \textbf{2. Cloth-Simplified} : The cloth consists of an 81 joint composite that is a $9\times 9 $ grid. The robot needs to pick the corner joint of the cloth and move that to the target place. The action space is similar to the rope environment except the picking location can only be one of the four corners. In this environment, the goal is to flatten the cloth in the middle of the table. Our reward function is the intersection of the binary mask of the cloth with the goal cloth configuration. 

  \par \textbf{3. Cloth} :  In contrast to the \textit{Cloth-Simplified} environment that can only pick one of the 4 corners, \textit{Cloth} allows picking any point in the pixel of cloth (if it is trained with image observation) or any composite particle (if state observation is used). The reward used is the same as in \textit{Cloth-Simplified}. For both the Cloth and Cloth-Simplified environments, the cloth is initialized by applying a random action for the first 130 timesteps of each episode. In MuJoCo, the skin of the cloth can be simulated by uploading an image taken of a real cloth texture. In the domain randomization experiments, we randomize the cloth by switching out different textures.

\subsection{Learning Methods for Comparison}
To understand the significance of our algorithm, we compare the following learning methods: random, independent, conditional, learned placing with uniform pick, and MVP (ours) as described below.

\begin{itemize}
    \item \textbf{Random}: We sample pick actions uniformly over available pick locations and place actions uniformly over the action space of the robot.
    
    \item \textbf{Independent / Joint}: We use a joint factorization of $p(a_{pick}, a_{place}|o)$ by simultaneously outputting the $a_{pick}$ and $a_{place}$. Alternatively, we label it as Independent to distinguish it from the Conditional baseline.
    
    \item \textbf{Conditional}: We first choose a pick location, and then choose a place vector distance given the pick location, modeled as $p(a_{pick}|o)\times p(a_{place}|a_{pick},o)$.
    
    \item \textbf{Learned Placing with Uniform Pick}: We use the conditional distribution $p(a_{place}|a_{pick},o)$, where $a_{pick}$ is uniformly sampled from available pick locations.
    
    \item \textbf{MVP (ours)}: We use the trained learned placing with uniform pick policy and choose $a_{pick}$ by maximizing over the learned Q-function.
\end{itemize}

Our experimental results for various model architectures on the rope and cloth environments are shown in ~\Figref{fig:sim_plots_state} and \Figref{fig:sim_plots_image}. We trained the Independent and Conditional baselines using a modified environment, where an extra positive reward is given for successfully outputting a pick location on the cloth or rope. This is to allow a more fair comparison with the Random, Learned Placing with Uniform Pick, and MVP (ours) policies which have prior access to the image segmentations.

\subsection{Training Details}
For the training in the simulation, we use SAC~\cite{haarnoja2018soft} as our off-policy algorithm and make a few modifications on the rlpyt code-base~\cite{stooke2019rlpyt}. For state-based experiments, we use an MLP with 2 hidden layers of 256 units each; approximately 150k parameters. For image-based experiments, we use a CNN with 3 convolutional layers with channel sizes 64, 64, and 4, accordingly, and each with a kernel size of 3 and a stride of 2. This is followed by with 2 fully connected hidden layers of 256 units each. In total approximately 200k parameters are learned. For all models, we repeat the pick information 50 times before concatenating with the state observations or flattened image embeddings so that the pick information and the observation embeddings are weighted equally, which improves performance. The horizon for Rope is 200 and 120 for both Cloth environments. The minimum replay pool size is 2000 for Rope and 1200 for the Cloth environments. The image size used for all environments is $64 \times 64 \times 3$. We perform parallel environment sampling to speed-up overall training by 3$-$5 times. For both rope and cloth experiments, the total compute time on one TitanX GPU and 4 CPU cores is roughly 4-6 hours. In the case of rope experiments, a reasonable policy can be obtained in one sixth of the training time, and about a half of the training time for cloth experiments. All of our training code, baselines, and simulation environments will be publicly released.

\subsection{Does conditional pick-place learning help?}
\label{sec:fact_benefits}
To understand the effects of our learning technique, we compare our learned placing with uniform pick technique with the independent representation in \Figref{fig:sim_plots_state}. We can see that using our proposed method shows improvement in learning speed for state-based cloth experiments, and image-based experiments in general. The state-based rope experiments do not show much of a difference due to the inherent simplicity of the tasks. Our method shows significantly higher rewards in the cloth simplified environment, and learns about 2X faster in the harder cloth environment. For image-based experiments, the baseline methods do no better than random while our method gives an order of magnitude (5-10X) higher performance for reward reached. The independent and conditional factored policies for image-based cloth spreading end up performing worse than random, suggesting mode collapse that commonly occurs in difficult optimization problems~\cite{goodfellow2014generative}. Note that to strengthen the baselines, we add additional rewards to bias the pick points on the cloth; However, this still does not significantly improve performance for the challenging image based tasks. This demonstrates that conditional learning indeed speeds up learning for deformable object manipulation especially when the observation is an image.


\subsection{Does setting the picking policy based on MVP help?}
One of the key contributions of this work is to use the placing value to inform the picking policy (\Eqref{eq:mvp}) without explicitly training the picking policy. As we see in both state-based (\Figref{fig:sim_plots_state}) and image-based case (\Figref{fig:sim_plots_image}) training with MVP gives consistently better performance. Even when our conditional policies with uniform pick location fall below the baselines as seen in Cloth (State) and Rope (State), using MVP significantly improves the performance. Note that although MVP brings relatively smaller boosts in performance compared to the gains brought by the learned placing with uniform pick method, we observe that the learned placing with uniform pick policy already achieves a high success rate on completing the task, and even a small boost in performance is visually substantial when running evaluations in simulation and on our real robot.

\subsection{How do we transfer our policies to a real robot?}

\begin{figure}[t!]
  \begin{center}
    \includegraphics[width = 0.9\linewidth]{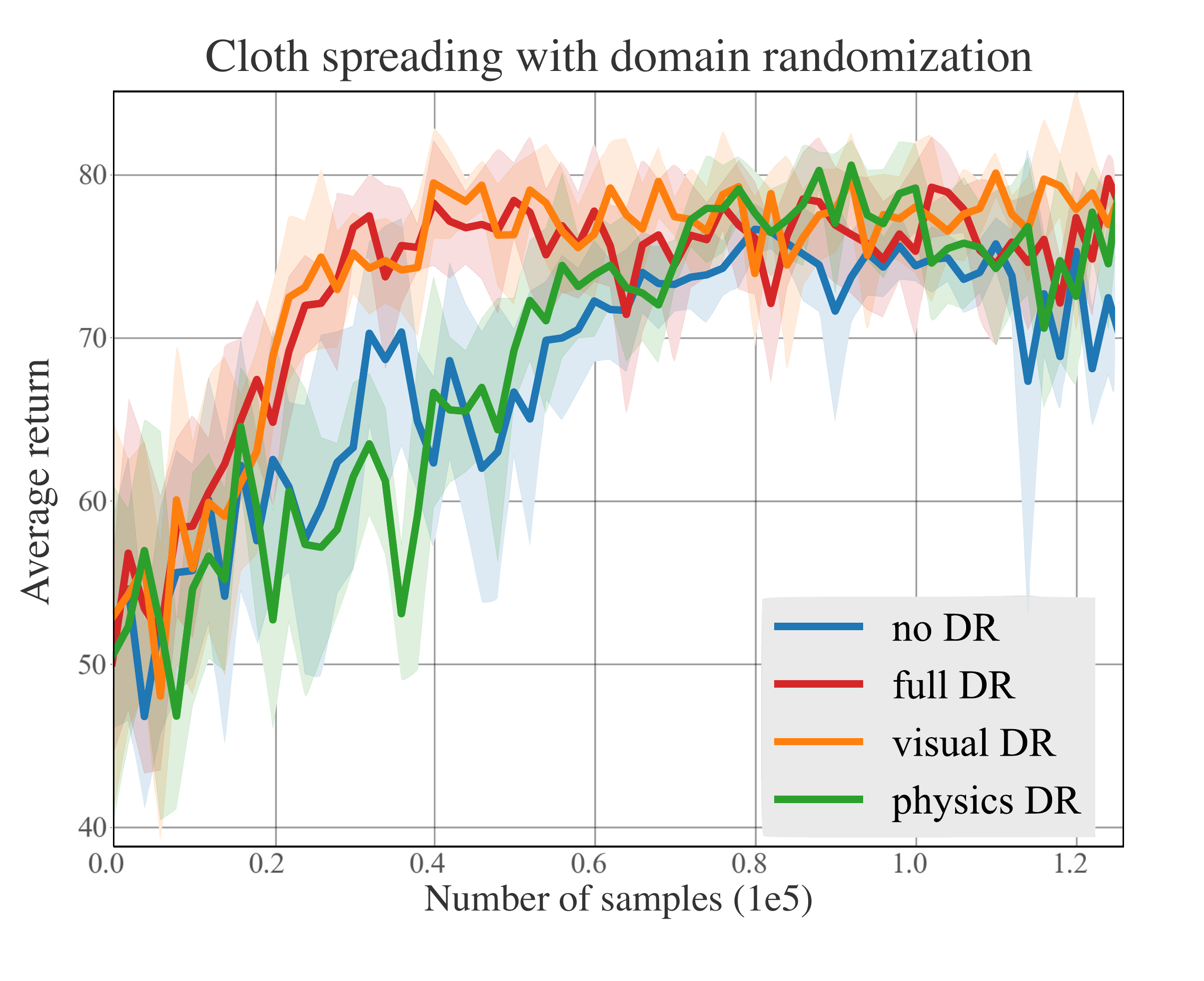}
  \end{center}
  \caption{Learning comparisons between different forms of domain randomization (DR) on cloth-spreading trained with MVP. This is evaluated in simulation across 5 random seeds and shaded with $\pm$ 1 standard deviation.}
\label{fig:DR}
\end{figure}

\begin{figure}[t!]
  \begin{center}
    \includegraphics[width = 1.0\linewidth]{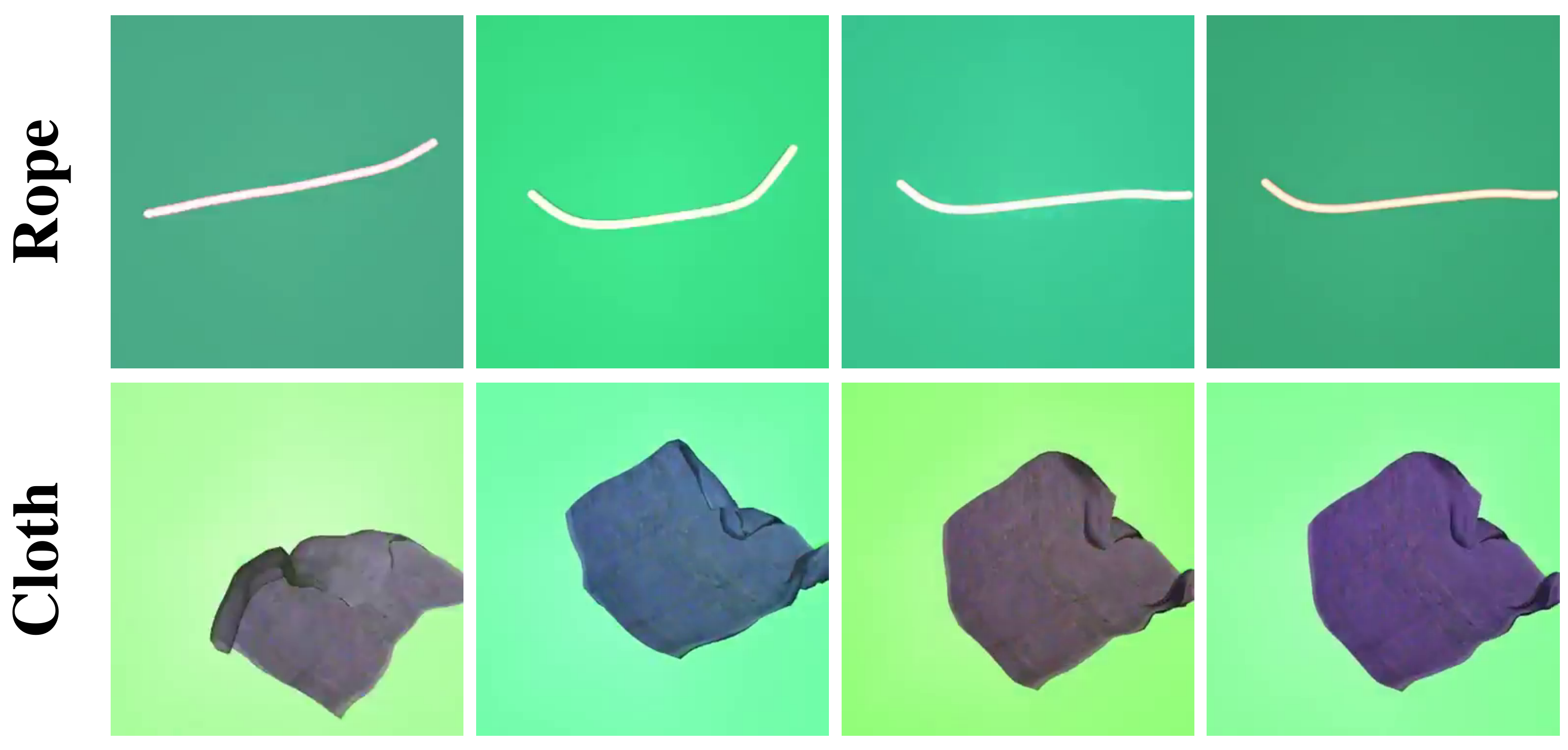}
  \end{center}
  \caption{Examples of domain randomization applied in the rope and cloth environments.}
\label{fig:DR_ex}
\end{figure}

To transfer our policies to the real-robot, we use domain randomization (DR)~\cite{tobin2018domain,pinto2017asymmetric,sadeghi2016cad2rl} in the simulator along with using images of real cloths. Randomization is performed on visual parameters (lighting and textures) as well physics (mass and joint friction) of the cloth. Examples of randomized observations can be seen in \Figref{fig:DR_ex}. Additionally, in simulation evaluation, we notice no degradation in performance due to DR while training using MVP as seen in \Figref{fig:DR}. 

In order to perform actions on our PR2 robot, we first calibrate pixel-space actions with robot actions. This is done by collecting 4-5 points mapping between robot $x,y$ coordinates to image row, column pixel locations, and fitting a simple linear map. Next, we capture RGB images from a head-mounted camera on our PR2 robot(\Figref{fig:intro}(a)) and input the image into our policy learned in the simulator. Since $a_{pick}$ and $a_{place}$ are both defined as points on the image, we can easily command the robot to perform pick-place operations on the deformable object placed on the green table by planning with Moveit!~\cite{chitta2012moveit}.


\subsection{Evaluation on the real robot}
We evaluate our policy on the \textit{rope-spread} and \textit{cloth-spread} experiments. As seen in \Figref{fig:robot_exp}, policies trained using MVP are successfully able to complete both spreading tasks. For our cloth spreading experiment, we also note that due to domain randomization, a single policy can spread cloths of different colors. For quantitative evaluations, we select 4 start configurations for the cloth and the rope and compare with various baselines (Table \ref{tab:robot_exp}) on the spread coverage metric. For the rope task, we run the policies for 20 steps, while for the much harder cloth task we run policies for 150 steps. The large gap between MVP trained policies and independent policies supports our hypothesis that the conditional structure is crucial for learning deformable object manipulation. Robot execution videos can be accessed from the video submission.

\begin{table}[h!]
\begin{center}
\begin{tabular}{ |c |c |c| c|c|c|}
\hline
  \multirow{3}{4em}{Domains}& 
  \multirow{3}{3em}{Random policy} & \multirow{3}{5em}{Conditional Pick-Place} & \multirow{3}{8em}{Independent / Joint policy} & 
  \multirow{3}{3em}{MVP (ours)} \\ 
  &&&& \\
 &&&& \\
\hline
 Rope & 0.34 & 0.16 & 0.21 & \textbf{0.48}\\  
 Cloth & 0.59 & 0.34 &0.32 & \textbf{0.84}\\
\hline
\end{tabular}
\end{center}
\caption{Average goal area intersection coverage for rope and cloth spreading tasks on the PR2 robot. 
}
\label{tab:robot_exp}
\end{table}


\section{Conclusion and Future Work}
We have proposed a conditional learning approach for learning to manipulating deformable objects. We have shown this significantly improves sample complexity. To our knowledge, this is the first work that trains RL from scratch for deformable object manipulation and demonstrates it on real robot. We believe this work can open up many exciting avenues for deformable object manipulation from bubble wrapping a rigid object to folding a T-shirt, which pose additional challenges in specifying a reward function and handling partial observability. Additionally, since our technique only assumes an actor-critic algorithm, we believe it can be combined with existing learning from demonstration based techniques to obtain further improvements in performance.


\section{Acknowledgements}
We thank AWS for computing resources and Boren Tsai for support in setting up the robot. We also gratefully acknowledge the support from Komatsu Ltd., The Open Philanthropy Project, Berkeley DeepDrive, NSF, and the ONR Pecase award.

\bibliographystyle{plainnat}
\bibliography{references}

\begin{thebibliography}{69}
\providecommand{\natexlab}[1]{#1}
\providecommand{\url}[1]{\texttt{#1}}
\expandafter\ifx\csname urlstyle\endcsname\relax
  \providecommand{\doi}[1]{doi: #1}\else
  \providecommand{\doi}{doi: \begingroup \urlstyle{rm}\Url}\fi

\bibitem[Andrychowicz et~al.(2018)Andrychowicz, Baker, Chociej, Jozefowicz,
  McGrew, Pachocki, Petron, Plappert, Powell, Ray,
  et~al.]{andrychowicz2018learning}
Marcin Andrychowicz, Bowen Baker, Maciek Chociej, Rafal Jozefowicz, Bob McGrew,
  Jakub Pachocki, Arthur Petron, Matthias Plappert, Glenn Powell, Alex Ray,
  et~al.
\newblock Learning dexterous in-hand manipulation.
\newblock \emph{arXiv preprint arXiv:1808.00177}, 2018.

\bibitem[Berenson(2013)]{berenson2013manipulation}
Dmitry Berenson.
\newblock Manipulation of deformable objects without modeling and simulating
  deformation.
\newblock In \emph{2013 IEEE/RSJ International Conference on Intelligent Robots
  and Systems}, pages 4525--4532. IEEE, 2013.

\bibitem[Brockman et~al.(2016)Brockman, Cheung, Pettersson, Schneider,
  Schulman, Tang, and Zaremba]{brockman2016openai}
Greg Brockman, Vicki Cheung, Ludwig Pettersson, Jonas Schneider, John Schulman,
  Jie Tang, and Wojciech Zaremba.
\newblock Openai gym.
\newblock \emph{arXiv preprint arXiv:1606.01540}, 2016.

\bibitem[Brooks(1983)]{brooks1983planning}
Rodney~A Brooks.
\newblock Planning collision-free motions for pick-and-place operations.
\newblock \emph{The International Journal of Robotics Research}, 2\penalty0
  (4):\penalty0 19--44, 1983.

\bibitem[Chitta et~al.(2012)Chitta, Sucan, and Cousins]{chitta2012moveit}
Sachin Chitta, Ioan Sucan, and Steve Cousins.
\newblock Moveit![ros topics].
\newblock \emph{IEEE Robotics \& Automation Magazine}, 19\penalty0
  (1):\penalty0 18--19, 2012.

\bibitem[Duan et~al.(2016)Duan, Chen, Houthooft, Schulman, and
  Abbeel]{duan2016benchmarking}
Yan Duan, Xi~Chen, Rein Houthooft, John Schulman, and Pieter Abbeel.
\newblock Benchmarking deep reinforcement learning for continuous control.
\newblock In \emph{International Conference on Machine Learning}, pages
  1329--1338, 2016.

\bibitem[Essahbi et~al.(2012)Essahbi, Bouzgarrou, and Gogu]{essahbi2012soft}
Nabil Essahbi, Belhassen~Chedli Bouzgarrou, and Grigore Gogu.
\newblock Soft material modeling for robotic manipulation.
\newblock In \emph{Applied Mechanics and Materials}, volume 162, pages
  184--193. Trans Tech Publ, 2012.

\bibitem[Finn and Levine(2017)]{finn2017deep}
Chelsea Finn and Sergey Levine.
\newblock Deep visual foresight for planning robot motion.
\newblock In \emph{2017 IEEE International Conference on Robotics and
  Automation (ICRA)}, pages 2786--2793. IEEE, 2017.

\bibitem[Frank et~al.(2011)Frank, Stachniss, Abdo, and
  Burgard]{frank2011efficient}
Barbara Frank, Cyrill Stachniss, Nichola Abdo, and Wolfram Burgard.
\newblock Efficient motion planning for manipulation robots in environments
  with deformable objects.
\newblock In \emph{2011 IEEE/RSJ International Conference on Intelligent Robots
  and Systems}, pages 2180--2185. IEEE, 2011.

\bibitem[{Fujimoto} et~al.(2018){Fujimoto}, {van Hoof}, and
  {Meger}]{2018arXiv180209477F}
Scott {Fujimoto}, Herke {van Hoof}, and David {Meger}.
\newblock {Addressing Function Approximation Error in Actor-Critic Methods}.
\newblock \emph{arXiv e-prints}, art. arXiv:1802.09477, Feb 2018.

\bibitem[Goodfellow et~al.(2014)Goodfellow, Pouget-Abadie, Mirza, Xu,
  Warde-Farley, Ozair, Courville, and Bengio]{goodfellow2014generative}
Ian Goodfellow, Jean Pouget-Abadie, Mehdi Mirza, Bing Xu, David Warde-Farley,
  Sherjil Ozair, Aaron Courville, and Yoshua Bengio.
\newblock Generative adversarial nets.
\newblock In \emph{NIPS}, 2014.

\bibitem[Gupta et~al.(2018)Gupta, Murali, Gandhi, and Pinto]{gupta2018robot}
Abhinav Gupta, Adithyavairavan Murali, Dhiraj~Prakashchand Gandhi, and Lerrel
  Pinto.
\newblock Robot learning in homes: Improving generalization and reducing
  dataset bias.
\newblock In \emph{Advances in Neural Information Processing Systems}, pages
  9094--9104, 2018.

\bibitem[{Haarnoja} et~al.(2018){Haarnoja}, {Zhou}, {Abbeel}, and
  {Levine}]{2018arXiv180101290H}
Tuomas {Haarnoja}, Aurick {Zhou}, Pieter {Abbeel}, and Sergey {Levine}.
\newblock {Soft Actor-Critic: Off-Policy Maximum Entropy Deep Reinforcement
  Learning with a Stochastic Actor}.
\newblock \emph{arXiv e-prints}, art. arXiv:1801.01290, Jan 2018.

\bibitem[Haarnoja et~al.(2018)Haarnoja, Zhou, Hartikainen, Tucker, Ha, Tan,
  Kumar, Zhu, Gupta, Abbeel, et~al.]{haarnoja2018soft}
Tuomas Haarnoja, Aurick Zhou, Kristian Hartikainen, George Tucker, Sehoon Ha,
  Jie Tan, Vikash Kumar, Henry Zhu, Abhishek Gupta, Pieter Abbeel, et~al.
\newblock Soft actor-critic algorithms and applications.
\newblock \emph{arXiv preprint arXiv:1812.05905}, 2018.

\bibitem[Henrich and W{\"o}rn(2012)]{henrich2012robot}
Dominik Henrich and Heinz W{\"o}rn.
\newblock \emph{Robot manipulation of deformable objects}.
\newblock Springer Science \& Business Media, 2012.

\bibitem[Howard and Bekey(2000)]{howard2000intelligent}
Ayanna~M Howard and George~A Bekey.
\newblock Intelligent learning for deformable object manipulation.
\newblock \emph{Autonomous Robots}, 9\penalty0 (1):\penalty0 51--58, 2000.

\bibitem[Hu et~al.(2018)Hu, Sun, and Pan]{hu2018three}
Zhe Hu, Peigen Sun, and Jia Pan.
\newblock Three-dimensional deformable object manipulation using fast online
  gaussian process regression.
\newblock \emph{IEEE Robotics and Automation Letters}, 3\penalty0 (2):\penalty0
  979--986, 2018.

\bibitem[Jia et~al.(2018)Jia, Hu, Pan, Manocha, and Pan]{jia2018learning}
Biao Jia, Zhe Hu, Zherong Pan, Dinesh Manocha, and Jia Pan.
\newblock Learning-based feedback controller for deformable object
  manipulation.
\newblock \emph{arXiv preprint arXiv:1806.09618}, 2018.

\bibitem[Jim{\'e}nez(2012)]{jimenez2012survey}
P~Jim{\'e}nez.
\newblock Survey on model-based manipulation planning of deformable objects.
\newblock \emph{Robotics and computer-integrated manufacturing}, 28\penalty0
  (2):\penalty0 154--163, 2012.

\bibitem[Johnson et~al.(2015)Johnson, Shrewsbury, Bertrand, Wu, Duran, Floyd,
  Abeles, Stephen, Mertins, Lesman, et~al.]{johnson2015team}
Matthew Johnson, Brandon Shrewsbury, Sylvain Bertrand, Tingfan Wu, Daniel
  Duran, Marshall Floyd, Peter Abeles, Douglas Stephen, Nathan Mertins, Alex
  Lesman, et~al.
\newblock Team ihmc's lessons learned from the darpa robotics challenge trials.
\newblock \emph{Journal of Field Robotics}, 32\penalty0 (2):\penalty0 192--208,
  2015.

\bibitem[Kaelbling et~al.(1996)Kaelbling, Littman, and
  Moore]{kaelbling1996reinforcement}
Leslie~Pack Kaelbling, Michael~L Littman, and Andrew~W Moore.
\newblock Reinforcement learning: A survey.
\newblock \emph{Journal of artificial intelligence research}, 4:\penalty0
  237--285, 1996.

\bibitem[Kakade(2002)]{kakade2002natural}
Sham~M Kakade.
\newblock A natural policy gradient.
\newblock In \emph{Advances in neural information processing systems}, pages
  1531--1538, 2002.

\bibitem[Khalil and Payeur(2010)]{khalil2010dexterous}
Fouad~F Khalil and Pierre Payeur.
\newblock Dexterous robotic manipulation of deformable objects with
  multi-sensory feedback-a review.
\newblock In \emph{Robot Manipulators Trends and Development}. IntechOpen,
  2010.

\bibitem[Kurutach et~al.(2018)Kurutach, Clavera, Duan, Tamar, and
  Abbeel]{kurutach2018model}
Thanard Kurutach, Ignasi Clavera, Yan Duan, Aviv Tamar, and Pieter Abbeel.
\newblock Model-ensemble trust-region policy optimization.
\newblock \emph{arXiv preprint arXiv:1802.10592}, 2018.

\bibitem[Lee et~al.(2015)Lee, Gupta, Lu, Levine, and Abbeel]{lee2015learning}
Alex~X Lee, Abhishek Gupta, Henry Lu, Sergey Levine, and Pieter Abbeel.
\newblock Learning from multiple demonstrations using trajectory-aware
  non-rigid registration with applications to deformable object manipulation.
\newblock In \emph{2015 IEEE/RSJ International Conference on Intelligent Robots
  and Systems (IROS)}, pages 5265--5272. IEEE, 2015.

\bibitem[Levine et~al.(2016{\natexlab{a}})Levine, Finn, Darrell, and
  Abbeel]{levine2016end}
Sergey Levine, Chelsea Finn, Trevor Darrell, and Pieter Abbeel.
\newblock End-to-end training of deep visuomotor policies.
\newblock \emph{JMLR}, 2016{\natexlab{a}}.

\bibitem[Levine et~al.(2016{\natexlab{b}})Levine, Pastor, Krizhevsky, and
  Quillen]{levine2016learning}
Sergey Levine, Peter Pastor, Alex Krizhevsky, and Deirdre Quillen.
\newblock Learning hand-eye coordination for robotic grasping with deep
  learning and large-scale data collection.
\newblock \emph{ISER}, 2016{\natexlab{b}}.

\bibitem[{Lillicrap} et~al.(2015){Lillicrap}, {Hunt}, {Pritzel}, {Heess},
  {Erez}, {Tassa}, {Silver}, and {Wierstra}]{2015arXiv150902971L}
Timothy~P. {Lillicrap}, Jonathan~J. {Hunt}, Alexand~er {Pritzel}, Nicolas
  {Heess}, Tom {Erez}, Yuval {Tassa}, David {Silver}, and Daan {Wierstra}.
\newblock {Continuous control with deep reinforcement learning}.
\newblock \emph{arXiv e-prints arXiv:1509.02971}, 2015.

\bibitem[Lillicrap et~al.(2015)Lillicrap, Hunt, Pritzel, Heess, Erez, Tassa,
  Silver, and Wierstra]{lillicrap2015continuous}
Timothy~P Lillicrap, Jonathan~J Hunt, Alexander Pritzel, Nicolas Heess, Tom
  Erez, Yuval Tassa, David Silver, and Daan Wierstra.
\newblock Continuous control with deep reinforcement learning.
\newblock \emph{arXiv preprint arXiv:1509.02971}, 2015.

\bibitem[Lozano-P{\'e}rez et~al.(1989)Lozano-P{\'e}rez, Jones, Mazer, and
  O'Donnell]{lozano1989task}
Tom{\'a}s Lozano-P{\'e}rez, Joseph~L. Jones, Emmanuel Mazer, and Patrick~A.
  O'Donnell.
\newblock Task-level planning of pick-and-place robot motions.
\newblock \emph{Computer}, 22\penalty0 (3):\penalty0 21--29, 1989.

\bibitem[Mahler et~al.(2016)Mahler, Pokorny, Hou, Roderick, Laskey, Aubry,
  Kohlhoff, Kröger, Kuffner, and Goldberg]{mahler2016dexnet}
J.~Mahler, F.~T. Pokorny, B.~Hou, M.~Roderick, M.~Laskey, M.~Aubry,
  K.~Kohlhoff, T.~Kröger, J.~Kuffner, and K.~Goldberg.
\newblock Dex-net 1.0: A cloud-based network of 3d objects for robust grasp
  planning using a multi-armed bandit model with correlated rewards.
\newblock In \emph{ICRA}, 2016.

\bibitem[Maitin-Shepard et~al.(2010)Maitin-Shepard, Cusumano-Towner, Lei, and
  Abbeel]{maitin2010cloth}
Jeremy Maitin-Shepard, Marco Cusumano-Towner, Jinna Lei, and Pieter Abbeel.
\newblock Cloth grasp point detection based on multiple-view geometric cues
  with application to robotic towel folding.
\newblock In \emph{2010 IEEE International Conference on Robotics and
  Automation}, pages 2308--2315. IEEE, 2010.

\bibitem[Matas et~al.(2018)Matas, James, and Davison]{matas2018sim}
Jan Matas, Stephen James, and Andrew~J Davison.
\newblock Sim-to-real reinforcement learning for deformable object
  manipulation.
\newblock \emph{arXiv preprint arXiv:1806.07851}, 2018.

\bibitem[Mayer et~al.(2008)Mayer, Gomez, Wierstra, Nagy, Knoll, and
  Schmidhuber]{mayer2008system}
Hermann Mayer, Faustino Gomez, Daan Wierstra, Istvan Nagy, Alois Knoll, and
  J{\"u}rgen Schmidhuber.
\newblock A system for robotic heart surgery that learns to tie knots using
  recurrent neural networks.
\newblock \emph{Advanced Robotics}, 22\penalty0 (13-14):\penalty0 1521--1537,
  2008.

\bibitem[McConachie and Berenson(2018)]{mcconachie2018estimating}
Dale McConachie and Dmitry Berenson.
\newblock Estimating model utility for deformable object manipulation using
  multiarmed bandit methods.
\newblock \emph{IEEE Transactions on Automation Science and Engineering},
  15\penalty0 (3):\penalty0 967--979, 2018.

\bibitem[McConachie et~al.(2017)McConachie, Ruan, and
  Berenson]{mcconachie2017interleaving}
Dale McConachie, Mengyao Ruan, and Dmitry Berenson.
\newblock Interleaving planning and control for deformable object manipulation.
\newblock In \emph{International Symposium on Robotics Research (ISRR)}, 2017.

\bibitem[Mnih et~al.(2015)Mnih, Kavukcuoglu, Silver, Rusu, Veness, Bellemare,
  Graves, Riedmiller, Fidjeland, Ostrovski, et~al.]{mnih2015human}
Volodymyr Mnih, Koray Kavukcuoglu, David Silver, Andrei~A Rusu, Joel Veness,
  Marc~G Bellemare, Alex Graves, Martin Riedmiller, Andreas~K Fidjeland, Georg
  Ostrovski, et~al.
\newblock Human-level control through deep reinforcement learning.
\newblock \emph{Nature}, 518\penalty0 (7540):\penalty0 529, 2015.

\bibitem[{Mnih} et~al.(2016){Mnih}, {Puigdom{\`e}nech Badia}, {Mirza},
  {Graves}, {Lillicrap}, {Harley}, {Silver}, and
  {Kavukcuoglu}]{2016arXiv160201783M}
Volodymyr {Mnih}, Adri{\`a} {Puigdom{\`e}nech Badia}, Mehdi {Mirza}, Alex
  {Graves}, Timothy~P. {Lillicrap}, Tim {Harley}, David {Silver}, and Koray
  {Kavukcuoglu}.
\newblock {Asynchronous Methods for Deep Reinforcement Learning}.
\newblock \emph{arXiv e-prints}, art. arXiv:1602.01783, Feb 2016.

\bibitem[Moll and Kavraki(2006)]{moll2006path}
Mark Moll and Lydia~E Kavraki.
\newblock Path planning for deformable linear objects.
\newblock \emph{IEEE Transactions on Robotics}, 22\penalty0 (4):\penalty0
  625--636, 2006.

\bibitem[Murali et~al.(2018)Murali, Pinto, Gandhi, and Gupta]{murali2018cassl}
Adithyavairavan Murali, Lerrel Pinto, Dhiraj Gandhi, and Abhinav Gupta.
\newblock Cassl: Curriculum accelerated self-supervised learning.
\newblock In \emph{2018 IEEE International Conference on Robotics and
  Automation (ICRA)}, pages 6453--6460. IEEE, 2018.

\bibitem[Nair et~al.(2017)Nair, Chen, Agrawal, Isola, Abbeel, Malik, and
  Levine]{nair2017combining}
Ashvin Nair, Dian Chen, Pulkit Agrawal, Phillip Isola, Pieter Abbeel, Jitendra
  Malik, and Sergey Levine.
\newblock Combining self-supervised learning and imitation for vision-based
  rope manipulation.
\newblock In \emph{2017 IEEE International Conference on Robotics and
  Automation (ICRA)}, pages 2146--2153. IEEE, 2017.

\bibitem[Navarro-Alarcon et~al.(2014)Navarro-Alarcon, Liu, Romero, and
  Li]{navarro2014visual}
David Navarro-Alarcon, Yun-hui Liu, Jose~Guadalupe Romero, and Peng Li.
\newblock On the visual deformation servoing of compliant objects: Uncalibrated
  control methods and experiments.
\newblock \emph{The International Journal of Robotics Research}, 33\penalty0
  (11):\penalty0 1462--1480, 2014.

\bibitem[Piera{\'n}ski et~al.(2001)Piera{\'n}ski, Przyby{\l}, and
  Stasiak]{pieranski2001tight}
Piotr Piera{\'n}ski, Sylwester Przyby{\l}, and Andrzej Stasiak.
\newblock Tight open knots.
\newblock \emph{The European Physical Journal E}, 6\penalty0 (2):\penalty0
  123--128, 2001.

\bibitem[Pinto and Gupta(2016)]{pinto2016supersizing}
Lerrel Pinto and Abhinav Gupta.
\newblock Supersizing self-supervision: Learning to grasp from 50k tries and
  700 robot hours.
\newblock \emph{ICRA}, 2016.

\bibitem[Pinto et~al.(2017)Pinto, Andrychowicz, Welinder, Zaremba, and
  Abbeel]{pinto2017asymmetric}
Lerrel Pinto, Marcin Andrychowicz, Peter Welinder, Wojciech Zaremba, and Pieter
  Abbeel.
\newblock Asymmetric actor critic for image-based robot learning.
\newblock \emph{arXiv preprint arXiv:1710.06542}, 2017.

\bibitem[Rajeswaran et~al.(2017)Rajeswaran, Kumar, Gupta, Vezzani, Schulman,
  Todorov, and Levine]{rajeswaran2017learning}
Aravind Rajeswaran, Vikash Kumar, Abhishek Gupta, Giulia Vezzani, John
  Schulman, Emanuel Todorov, and Sergey Levine.
\newblock Learning complex dexterous manipulation with deep reinforcement
  learning and demonstrations.
\newblock \emph{arXiv preprint arXiv:1709.10087}, 2017.

\bibitem[Rodriguez et~al.(2006)Rodriguez, Tang, Lien, and
  Amato]{rodriguez2006obstacle}
Samuel Rodriguez, Xinyu Tang, Jyh-Ming Lien, and Nancy~M Amato.
\newblock An obstacle-based rapidly-exploring random tree.
\newblock In \emph{Proceedings 2006 IEEE International Conference on Robotics
  and Automation, 2006. ICRA 2006.}, pages 895--900. IEEE, 2006.

\bibitem[Sadeghi and Levine(2016)]{sadeghi2016cad2rl}
Fereshteh Sadeghi and Sergey Levine.
\newblock Cad2rl: Real single-image flight without a single real image.
\newblock \emph{arXiv preprint arXiv:1611.04201}, 2016.

\bibitem[Saha and Isto(2007)]{saha2007manipulation}
Mitul Saha and Pekka Isto.
\newblock Manipulation planning for deformable linear objects.
\newblock \emph{IEEE Transactions on Robotics}, 23\penalty0 (6):\penalty0
  1141--1150, 2007.

\bibitem[Schulman et~al.(2013{\natexlab{a}})Schulman, Gupta, Venkatesan,
  Tayson-Frederick, and Abbeel]{schulman2013case}
John Schulman, Ankush Gupta, Sibi Venkatesan, Mallory Tayson-Frederick, and
  Pieter Abbeel.
\newblock A case study of trajectory transfer through non-rigid registration
  for a simplified suturing scenario.
\newblock In \emph{2013 IEEE/RSJ International Conference on Intelligent Robots
  and Systems}, pages 4111--4117. IEEE, 2013{\natexlab{a}}.

\bibitem[Schulman et~al.(2013{\natexlab{b}})Schulman, Ho, Lee, and
  Abbeel]{schulman2013generalization}
John Schulman, Jonathan Ho, Cameron Lee, and Pieter Abbeel.
\newblock Generalization in robotic manipulation through the use of non-rigid
  registration.
\newblock In \emph{Proceedings of the 16th International Symposium on Robotics
  Research (ISRR)}, 2013{\natexlab{b}}.

\bibitem[Schulman et~al.(2013{\natexlab{c}})Schulman, Lee, Ho, and
  Abbeel]{schulman2013tracking}
John Schulman, Alex Lee, Jonathan Ho, and Pieter Abbeel.
\newblock Tracking deformable objects with point clouds.
\newblock In \emph{2013 IEEE International Conference on Robotics and
  Automation}, pages 1130--1137. IEEE, 2013{\natexlab{c}}.

\bibitem[Schulman et~al.(2015)Schulman, Levine, Abbeel, Jordan, and
  Moritz]{schulman2015trust}
John Schulman, Sergey Levine, Pieter Abbeel, Michael~I Jordan, and Philipp
  Moritz.
\newblock Trust region policy optimization.
\newblock In \emph{ICML}, pages 1889--1897, 2015.

\bibitem[Seita et~al.(2018)Seita, Jamali, Laskey, Tanwani, Berenstein,
  Baskaran, Iba, Canny, and Goldberg]{seita2018deep}
Daniel Seita, Nawid Jamali, Michael Laskey, Ajay~Kumar Tanwani, Ron Berenstein,
  Prakash Baskaran, Soshi Iba, John Canny, and Ken Goldberg.
\newblock Deep transfer learning of pick points on fabric for robot bed-making.
\newblock \emph{arXiv preprint arXiv:1809.09810}, 2018.

\bibitem[Seita et~al.(2019)Seita, Ganapathi, Hoque, Hwang, Cen, Tanwani,
  Balakrishna, Thananjeyan, Ichnowski, Jamali, Yamane, Iba, Canny, and
  Goldberg]{seita2019deep}
Daniel Seita, Aditya Ganapathi, Ryan Hoque, Minho Hwang, Edward Cen, Ajay~Kumar
  Tanwani, Ashwin Balakrishna, Brijen Thananjeyan, Jeffrey Ichnowski, Nawid
  Jamali, Katsu Yamane, Soshi Iba, John Canny, and Ken Goldberg.
\newblock Deep imitation learning of sequential fabric smoothing policies.
\newblock \emph{arXiv preprint arXiv:1910.04854}, 2019.

\bibitem[Shimoga(1996)]{shimoga1996robot}
Karun~B Shimoga.
\newblock Robot grasp synthesis algorithms: A survey.
\newblock \emph{The International Journal of Robotics Research}, 15\penalty0
  (3):\penalty0 230--266, 1996.

\bibitem[Smolen and Patriciu(2009)]{smolen2009deformation}
Jerzy Smolen and Alexandru Patriciu.
\newblock Deformation planning for robotic soft tissue manipulation.
\newblock In \emph{2009 Second International Conferences on Advances in
  Computer-Human Interactions}, pages 199--204. IEEE, 2009.

\bibitem[Stooke and Abbeel(2019)]{stooke2019rlpyt}
Adam Stooke and Pieter Abbeel.
\newblock rlpyt: A research code base for deep reinforcement learning in
  pytorch.
\newblock \emph{arXiv preprint arXiv:1909.01500}, 2019.

\bibitem[Stria et~al.(2014)Stria, Prusa, Hlavac, Wagner, Petrik, Krsek, and
  Smutny]{stria2014garment}
Jan Stria, Daniel Prusa, Vaclav Hlavac, Libor Wagner, Vladimir Petrik, Pavel
  Krsek, and Vladimir Smutny.
\newblock Garment perception and its folding using a dual-arm robot.
\newblock In \emph{2014 IEEE/RSJ International Conference on Intelligent Robots
  and Systems}, pages 61--67. IEEE, 2014.

\bibitem[Sutton et~al.(1998)Sutton, Barto, et~al.]{sutton1998introduction}
Richard~S Sutton, Andrew~G Barto, et~al.
\newblock \emph{Introduction to reinforcement learning}, volume~2.
\newblock MIT press Cambridge, 1998.

\bibitem[Tassa et~al.(2018)Tassa, Doron, Muldal, Erez, Li, Casas, Budden,
  Abdolmaleki, Merel, Lefrancq, et~al.]{tassa2018deepmind}
Yuval Tassa, Yotam Doron, Alistair Muldal, Tom Erez, Yazhe Li, Diego de~Las
  Casas, David Budden, Abbas Abdolmaleki, Josh Merel, Andrew Lefrancq, et~al.
\newblock Deepmind control suite.
\newblock \emph{arXiv preprint arXiv:1801.00690}, 2018.

\bibitem[Tobin et~al.(2018)Tobin, Biewald, Duan, Andrychowicz, Handa, Kumar,
  McGrew, Ray, Schneider, Welinder, et~al.]{tobin2018domain}
Josh Tobin, Lukas Biewald, Rocky Duan, Marcin Andrychowicz, Ankur Handa, Vikash
  Kumar, Bob McGrew, Alex Ray, Jonas Schneider, Peter Welinder, et~al.
\newblock Domain randomization and generative models for robotic grasping.
\newblock In \emph{2018 IEEE/RSJ International Conference on Intelligent Robots
  and Systems (IROS)}, pages 3482--3489. IEEE, 2018.

\bibitem[Todorov et~al.(2012)Todorov, Erez, and Tassa]{todorov2012mujoco}
Emanuel Todorov, Tom Erez, and Yuval Tassa.
\newblock Mujoco: A physics engine for model-based control.
\newblock In \emph{2012 IEEE/RSJ International Conference on Intelligent Robots
  and Systems}, pages 5026--5033. IEEE, 2012.

\bibitem[Van~den Oord et~al.(2016)Van~den Oord, Kalchbrenner, Espeholt,
  Vinyals, Graves, et~al.]{van2016conditional}
Aaron Van~den Oord, Nal Kalchbrenner, Lasse Espeholt, Oriol Vinyals, Alex
  Graves, et~al.
\newblock Conditional image generation with pixelcnn decoders.
\newblock In \emph{Advances in neural information processing systems}, pages
  4790--4798, 2016.

\bibitem[Wada et~al.(2001)Wada, Hirai, Kawamura, and Kamiji]{wada2001robust}
Takahiro Wada, Shinichi Hirai, Sadao Kawamura, and Norimasa Kamiji.
\newblock Robust manipulation of deformable objects by a simple pid feedback.
\newblock In \emph{Proceedings 2001 ICRA. IEEE International Conference on
  Robotics and Automation (Cat. No. 01CH37164)}, volume~1, pages 85--90. IEEE,
  2001.

\bibitem[Wakamatsu et~al.(2006)Wakamatsu, Arai, and
  Hirai]{wakamatsu2006knotting}
Hidefumi Wakamatsu, Eiji Arai, and Shinichi Hirai.
\newblock Knotting/unknotting manipulation of deformable linear objects.
\newblock \emph{The International Journal of Robotics Research}, 25\penalty0
  (4):\penalty0 371--395, 2006.

\bibitem[Wang et~al.(2019)Wang, Kurutach, Liu, Abbeel, and
  Tamar]{wang2019learning}
Angelina Wang, Thanard Kurutach, Kara Liu, Pieter Abbeel, and Aviv Tamar.
\newblock Learning robotic manipulation through visual planning and acting.
\newblock \emph{arXiv preprint arXiv:1905.04411}, 2019.

\bibitem[Williams(1992)]{williams1992simple}
Ronald~J Williams.
\newblock Simple statistical gradient-following algorithms for connectionist
  reinforcement learning.
\newblock \emph{Machine learning}, 8\penalty0 (3-4):\penalty0 229--256, 1992.

\bibitem[Yousef et~al.(2011)Yousef, Boukallel, and
  Althoefer]{yousef2011tactile}
Hanna Yousef, Mehdi Boukallel, and Kaspar Althoefer.
\newblock Tactile sensing for dexterous in-hand manipulation in robotics—a
  review.
\newblock \emph{Sensors and Actuators A: physical}, 167\penalty0 (2):\penalty0
  171--187, 2011.

\end{thebibliography}

\end{document}